\documentclass[conference]{IEEEtran}
\IEEEoverridecommandlockouts
\usepackage{amsmath,amssymb,amsfonts}
\usepackage{algorithmic}
\usepackage{subfig}
\usepackage{tipa}
\usepackage{graphicx}
\usepackage{amsmath}
\usepackage{amsfonts}
\usepackage{algorithmic}
\usepackage{subfig}
\usepackage{graphicx}
\usepackage{textcomp}
\usepackage{xcolor}
\usepackage{tipa}
\usepackage{csquotes}

\usepackage{csquotes}
\usepackage{hyperref}
\usepackage{textcomp}
\usepackage{xcolor}
\def\BibTeX{{\rm B\kern-.05em{\sc i\kern-.025em b}\kern-.08em
    T\kern-.1667em\lower.7ex\hbox{E}\kern-.125emX}}
\begin{document}



\title{Leveraging Visemes for Better Visual Speech Representation and Lip Reading}

\author{\IEEEauthorblockN{1\textsuperscript{st} Javad Peymanfard}
\IEEEauthorblockA{\textit{Iran University of Science and Technology}\\
javad\_peymanfard@comp.iust.ac.ir}
\and
\IEEEauthorblockN{2\textsuperscript{nd} Vahid Saeedi}
\IEEEauthorblockA{\textit{Iran University of Science and Technology} \\
vahid\_saeedi@alumni.iust.ac.ir}
\and
\IEEEauthorblockN{3\textsuperscript{rd} Mohammad Reza Mohammadi}
\IEEEauthorblockA{\textit{Iran University of Science and Technology} \\
mrmohammadi@iust.ac.ir}
\and
\IEEEauthorblockN{4\textsuperscript{th} Hossein Zeinali}
\IEEEauthorblockA{\textit{Amirkabir University of Technology} \\
hzeinali@aut.ac.ir}
\and
\IEEEauthorblockN{5\textsuperscript{th} Nasser Mozayani}
\IEEEauthorblockA{\textit{Iran University of Science and Technology} \\
mozayani@iust.ac.ir}
}




\maketitle

\begin{abstract}
	Lip reading is a challenging task that has many potential applications in speech recognition, human-computer interaction, and security systems. However, existing lip reading systems often suffer from low accuracy due to the limitations of video features. In this paper, we propose a novel approach that leverages visemes, which are groups of phonetically similar lip shapes, to extract more discriminative and robust video features for lip reading. We evaluate our approach on various tasks, including word-level and sentence-level lip reading, and audio-visual speech recognition using the Arman-AV dataset, a large-scale Persian corpus. Our experimental results show that our viseme based approach consistently outperforms the state-of-the-art methods in all these tasks. The proposed method reduces the lip-reading word error rate (WER) by 9.1\% relative to the best previous method.
\end{abstract}


\section{Introduction}
\label{sec:intro}
%


Lip reading is a challenging task that requires deep learning models to achieve high accuracy. These models have surpassed traditional methods such as Hidden Markov models, but they need large-scale labeled datasets that cover the application domain. In lip reading, some methods directly extract the text from the input video with a single model, while others use a two-step approach \cite{hmm-speech, Peymanfard2022}. The two-step lip reading consists of extracting visemes from lip movements in the first step, and converting visemes into characters in the second step. Visemes are the visual equivalent of phonemes, but they do not have a one-to-one correspondence with phonemes. For example, the words "back" and "pack" have different phonemes, but the same lip movements, because the phonemes /\textipa{b}/ and /\textipa{p}/ correspond to one viseme. This issue makes lip reading, also known as visual speech recognition (VSR), more difficult than normal speech recognition.


Lip reading problems can be divided into two levels: word level and sentence level. At the word level, the problem is a classification task, where each visual speech sample belongs to a class labeled by the spoken word. However, as mentioned earlier, some different phonemes have the same visemes, which creates ambiguity and limits lip reading accuracy. For example, if several words correspond to one sequence of visemes, and the model predicts the most frequent word, the accuracy will reach the highest possible level (assuming that the model makes no mistake).


In sentence-level lip reading, we can use text processing methods such as language modeling to select the best candidate word (among the words that share the same viseme sequence) that fits the phrase context. Paper \cite{Peymanfard2022} proposes a two-step method that converts visemes into characters using a sequence-to-sequence model with an attention mechanism.

Speech recognition can be improved by using a combination of audio and visual inputs. AV-HuBERT~\cite{AV-HuBERT} is a well-known model in the field of audio-visual speech recognition, which is trained using a self-supervised learning method. AV-HuBERT recognizes the sentence spoken in the video at the character level by capturing both the video and the sound. The model is trained in two stages: first, it learns a general representation of the problem space that is not specific to any language or task; second, it fine-tunes the representation for the target language and task.

In this research, we used the pre-trained AV-HuBERT model and fine-tuned it with a novel approach. For the audio part, we followed the same procedure as AV-HuBERT; but for the visual part, we trained the model with the corresponding visemes. With this idea, we were able to achieve higher accuracy than the original fine-tuning process. We evaluate our method on both single-modal (VSR) and multi-modal audio-visual speech recognition (AVSR) tasks and show that it outperforms the original AV-HuBERT method in both cases. We also demonstrate that our method can recognize the spoken words from the lip movements of the speaker without using any audio data.

The rest of the paper is organized as follows. First, in the related work section, we review the existing research in the lip reading field and follow it by describing our approach for fine-tuning AV-HuBERT with visemes in the proposed method section. Then, we compare our method with similar works on various tasks in the experiments section. Finally, in conclusion, we summarize the main contributions and findings of this paper.

\section{Related Works}
\label{sec:rw}


Lip reading problems can be divided into two levels: word-level and sentence-level. In the word-level category, methods consider the problem as a classification task, where the input videos are padded to achieve the same number of frames. The model takes the video frames as input and predicts the corresponding word class as output. On the other hand, in the sentence-level category, the video samples contain multiple words, and the model outputs the predicted sentence as a sequence of words.


Lip reading methods can be classified into two types: one-step and two-step. The one-step methods predict the text directly from the input video, while the two-step methods use an intermediate representation of visemes. Paper~\cite{hmm-speech} proposed a two-step lip reading method using Hidden Markov Models (HMM). It uses two separate HMM models: one to extract visemes from the input video, and another to convert visemes into characters.


The authors of \cite{wlas} proposed a deep learning model for lip reading called WLAS. This model is the first one developed for in-the-wild conditions and is used for sentence-level problems. The WLAS model takes both video and audio inputs. It consists of three components: Watch (image sequence encoder), Listen (audio sequence encoder), and Spell (decoder). The input data is fed to the Watch and Listen components in separate time units. The Watch consists of a VGG~\cite{Simonyan2014} feature extraction network and a long short-term memory (LSTM) network to preserve the semantic relevance of the input sequence. Listen is similar to Watch, except that the input of the LSTM is the Mel-frequency cepstral coefficient (MFCC) features of the audio. The encoded features are then passed to Spell which consists of an LSTM module, two separate attention modules (one for image and one for sound), and a multilayer perceptron (MLP) module. The output of the MLP goes to Softmax, and a character is predicted for the current time unit. This model is trained on the LRS3 dataset~\cite{Afouras2018} and achieves better performance than its previous works.


The authors of \cite{hybridctc} introduced a hybrid Connectionist temporal classification (CTC) with attention architecture for decoding the audio-visual encoded embeddings. The encoder architecture consists of BiLSTM layers that take convolutional neural network (CNN) features for the visual part and Mel-spectrogram features for the audio part. The decoder architecture consists of CTC loss and attention modules. The paper explores two ways of combining audio and visual features: early fusion and late fusion. In early fusion, the encoder outputs one audio-visual embedding; in late fusion, the encoder outputs two separate embeddings for audio and image sequences. The paper shows that early fusion performs better than late one.


The paper \cite{davsr} compares two different loss functions for sentence-level audio-visual speech recognition: CTC loss and sequence-to-sequence loss. It proposes two different decoder architectures based on self-attention transformers, one for each loss function. The encoder architecture is the same for both models. It shows that the sequence-to-sequence model achieves better results than the CTC model for both audio-visual and visual speech recognition tasks.


AV-HuBERT~\cite{AV-HuBERT} is one of the state-of-the-art methods for lip reading, which is an end-to-end method for audio-visual speech recognition. This model takes both the audio and visual features of the speaker as input and outputs the corresponding text. AV-HuBERT is based on BERT~\cite{Devlin2018} and trains in a self-supervised way. In this model, the input video is split into several equal time intervals; then, the audio signal is processed by the MFCC network, and the video sequence is processed by the ResNet network with 3D convolution. The features of the audio and video hidden layers are combined and fed to the Transformer model. The model is first pre-trained with a large amount of unlabeled data, then fine-tuned with labeled samples. This improves the performance of the model significantly. For pre-training, similar to BERT, some input units are masked and the model is trained to predict the cluster of each masked unit. After that, the model is finetuned in a supervised way with the labeled part of the dataset.


The paper~\cite{cross-modal} proposed a knowledge distillation method for training lip-reading models. This method uses an ASR model as a teacher and a VSR model as a student. The teacher is a pre-trained model, and the student learns from the teacher's outputs. In this paper, the cross-entropy loss function is used to measure the error between the outputs of the VSR and ASR models, and the CTC loss function is used to measure the error between the text outputs of the VSR and ASR models.


In~\cite{Peymanfard2022}, a deep learning model was proposed for lip reading that uses a two-step approach. This model uses a 3D convolutional network to extract visemes from the lip movements in the video. Then, it uses a sequence-to-sequence model composed of a two-layer gated recurrent unit (GRU) network with an attention mechanism to convert the viseme sequence into a character sequence. Since some visemes can correspond to multiple phonemes, the attention mechanism can help to recognize the correct phoneme by relating the current and previous positions of the sequence.


There are various datasets available for lip reading. The LRW dataset~\cite{Chung2017} is one of the first datasets collected in natural conditions and on a large scale. It is a word-level dataset in English and consists of clips from UK BBC TV programs. The LRW-1000 dataset~\cite{Yang2019} is another large-scale dataset for the Chinese language. The paper~\cite{Peymanfard2022Dataset} proposed a word-level dataset for the Persian language, which contains 30 hours of clips from the Aparat website\footnote{\url{https://www.aparat.com/}} and more than 500 words. According to \cite{Peymanfard2022Dataset}, the dataset is collected under different audio and light conditions to cover various situations in real conditions.

\section{Proposed method}
\label{sec:method}

\subsection{Challenging Issue}
AV-HuBERT is a general model for audio and video representation. It uses a self-supervised technique to train the network with unlabeled data. This stage is called pre-training. At the end of this stage, the model can produce a suitable representation of the audio and visual information in a video. This information can include both the speaker and speech features. After this stage, the model needs to be fine-tuned for a specific task. For example, for recognition problems, such as visual speech recognition or lip reading, the information about the speaker is not relevant, and the speech in the video is the target. The AV-HuBERT paper conducts experiments on the largest available dataset for lip reading, called LRS3. These experiments involve fine-tuning the pre-trained model with different proportions of LRS3 labeled data. 
\subsection{Main Idea}
Since the viseme is the smallest meaningful unit in lip reading, our proposed technique is to use visemes to fine-tune AV-HuBERT. When we train the network for character-level lip reading, it learns not only the various lip movements of different speakers but also the language information from the dataset. We think that when we train a video representation model for lip reading, using a loss function based on character prediction, the representation implicitly contains language information from the dataset. Therefore, our proposed method is to train the network using the sequence transformation of characters to visemes and a loss function based on that. This method is expected to produce representations that have better information about lip movements, as well as higher generalization on other datasets, especially for samples with different languages.


In \cite{Peymanfard2022}, it has been shown that the conversion of visemes to characters can be done with high accuracy by using a large amount of textual data. Therefore, the main challenge for lip reading is to convert video to visemes, which can only be done by using labeled video data. In our proposed technique, we first obtain the phoneme sequence for each video using a semi-automatic technique and then transform it into a viseme sequence using a phoneme-to-viseme mapping. Then, we use these data to fine-tune the AV-HuBERT model. We expect that the video representation obtained by the model will be more suitable for lip reading than when the fine-tuning is done at the character level.


To compare these approaches, we can apply the fine-tuned models to lip reading problems in another language and compare the resulting accuracies. We have two models: one fine-tuned at the character level and the other at the viseme level. In general, we expect that a better video representation for lip reading will lead to a lower error rate. Therefore, by freezing the part of the model that is related to the representation and training the other layers of the lip reading network, we can obtain a more accurate model that uses better embedding. We can also use these two models for video representation in audio-visual speech recognition and compare their performance.

\section{Experiment and Results}
\label{sec:exp}

\subsection{Metrics}


In this section, we evaluate our proposed method, fine-tuning AV-HuBERT with visemes, by comparing it with the conventional fine-tuning method that uses the loss based on the final characters. All the experiments are done on Persian datasets for both word and sentence levels. We use different metrics for the evaluation: accuracy for the word-level task, and two error rates for the sentence-level tasks.


Accuracy is a metric that is used for classification problems. It measures the percentage of correct predictions out of all predictions. Since word-level lip reading is a classification task, we use accuracy to evaluate our method in this problem. We also calculate top-3, top-5, and top-10 accuracy for this problem. Top-N means that the model output is considered correct if the expected class is among the N highest probabilities.


In sentence-level problems, the input expressions can have different lengths and varieties. Therefore, we use Character Error Rate (CER) and Word Error Rate (WER) metrics that measure the error rate by comparing the ground truth and predicted word sequences. The CER metric algorithm transforms the predicted sequence into the ground-truth one by using the minimum number of deletions, insertions, and substitutions of characters. The CER value is computed by \eqref{CER}.

\begin{equation}
\label{CER}
CER = ((i_c + d_c + s_c) / n_c) * 100
\end{equation}


WER is another metric for comparing two sequences. WER counts the minimum number of word changes needed to obtain the ground-truth expression. In the same way as CER, it uses insertion, deletion, and substitution of words. WER is computed by \eqref{WER}.

\begin{equation}
\label{WER}
WER = ((i_w + d_w + s_w) / n_w) * 100
\end{equation}

\subsection{Datasets}


In this section, we give a brief overview of the datasets that we use to train, fine-tune and evaluate the speech recognition model with our method.


\textbf{LRS2}~\cite{wlas} is a lip-reading dataset collected from BBC channel. It is a public and large-scale in-the-wild dataset that has been used as a benchmark in several researches~\cite{lrs2user1, lrs2user2, lrs2user3}. We use this dataset to fine-tune the AV-HuBERT model. Note that the AV-HuBERT-base model has trained with the LRS3~\cite{Afouras2018} and VoxCeleb2~\cite{chung2018voxceleb2} datasets.


\textbf{Arman-AV}\cite{peymanfard2023multi} is a Persian dataset for audio-visual speech recognition. It is a large-scale in-the-wild dataset that contains over 220 hours of speech presented by 1760 celebrities. The dataset is collected from programs on the Aparat website. It also contains the speaker name as metadata, which makes it suitable for the speaker recognition task. We used the training set of Arman-AV for fine-tuning the model. We also evaluated all our methods on the test portion of this dataset.


\textbf{Common Voice}~\cite{commonvoice} is an audio-only speech dataset. It is a multi-lingual dataset that contains many languages (English, Chinese, Persian, etc). Mozilla created this dataset by collecting speech data using crowdsourcing. The used version of the dataset contains 2383 hours of speech in English and 352 hours in Persian. We used this dataset to train the audio part of the AVSR model.

\subsection{Word-level Visual Speech Recognition}


One of the challenges in lip reading is doing it at the word-level because we cannot use any cue from the surrounding words. In this task, the goal is to recognize the word uttered by the speaker from a short video clip (usually 100 milliseconds to 2 seconds) by extracting relevant visual information from the video frames. The words belong to a predefined vocabulary. This task can be considered as a video classification problem. There are suitable datasets for this task in different languages such as English, Chinese (Mandarin), and Persian. In this part, we evaluate and compare the quality of the representations obtained by our proposed method using the Persian dataset. The reason for not using English data, which is the most common data in this task, is to ensure a fair comparison of the two methods. In fact, in this part, we also examine and analyze the generalization ability of the representations learned by the two methods.


As shown in \autoref{tab:acc}, the model trained with our proposed method achieves higher accuracy on the test data. Other metrics for evaluating the quality of word-level lip reading, such as top-k accuracy, also indicate that the performance of our proposed model is better than the other models. These experiments demonstrate that the representation learned by the proposed model is more suitable for word-level lip reading.
\begin{figure*}[!t]
\centering
\includegraphics[width=\textwidth]{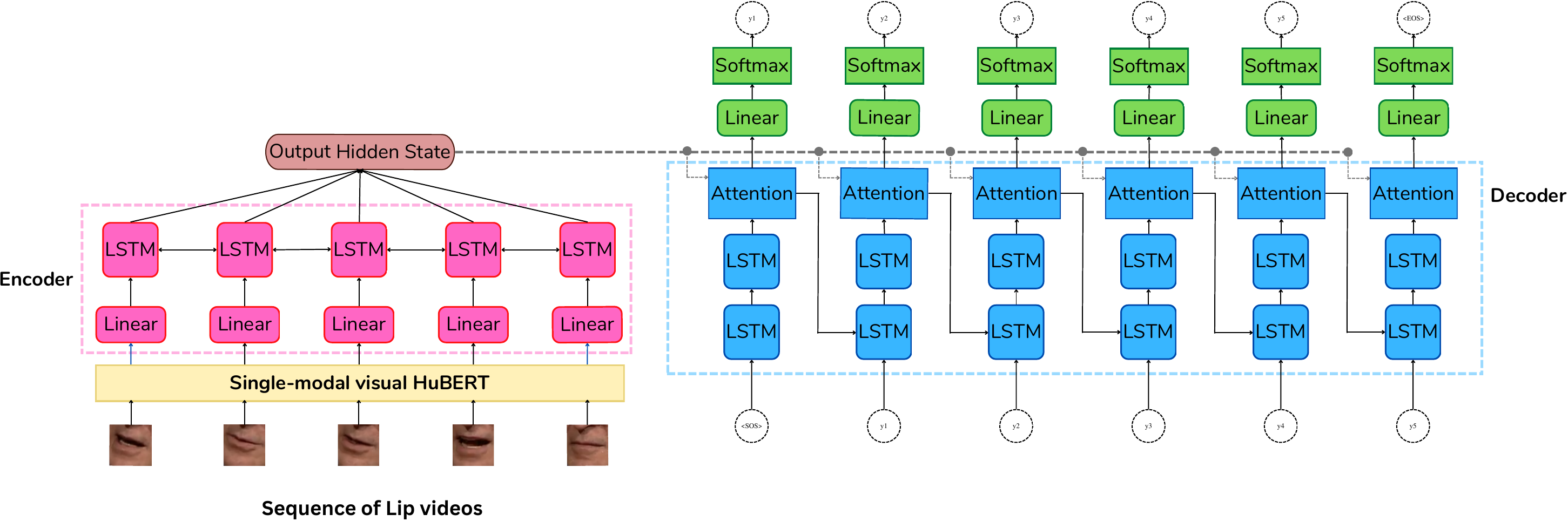}
\caption{The architecture used for the speech recognition task.}
\label{arch}
\end{figure*}

\subsection{Sentence-level Visual Speech Recognition}


The architecture of our method is shown in \autoref{arch}. A sequence of frames containing the speaker's lip is given to the single visual HuBERT modal, which outputs the visual feature vector. The visual features are fed into a decoder, consisting of a linear layer and a BiLSTM module, to obtain the hidden state vectors. The hidden states are passed through the decoder part, composed of two LSTM modules and an attention module, to output the probability vectors that are fed to a softmax layer to get the output visemes.


We evaluated our method on the Arman-AV dataset~\cite{peymanfard2023multi} for sentence-level speech recognition. The results in~\autoref{tab:vsr} show that fine-tuning the visual speech recognition model with visemes can reduce the uncertainty of the model, especially when different characters have similar lip movements. In fact, by using our proposed model, we obtained more suitable visual features for the lip reading task. We believe this is because the trained model has learned less language-specific information from the dataset (here English) and therefore has been able to extract more general features for other languages (Persian). According to~\autoref{tab:vsr} both CER and WER are significantly decreased by the proposed method.

\begin{table*}[btp!]
  \renewcommand{\arraystretch}{1.2}
  \begin{center}
    \caption{Sentence-level Visual Speech Recognition Error Rates in Percentage}
    \label{tab:vsr}
    \setlength\tabcolsep{12pt}
    \begin{tabular}{l | cc} 
      & \textbf{CER} & \textbf{WER} \\
      \textbf{Fine-tuned by Characters} & 49.54 & 77.22 \\ 
      \textbf{Fine-tuned by Visemes} & 44.08 & 70.19 \\
    \end{tabular}
  \end{center}
\end{table*}

\subsection{Audio-visual Speech Recognition}


In this section, we tackled the problem of audio-visual speech recognition. As we explained, this problem involves recognizing the speaker's speech using both audio and visual features. In this experiment, we used the model proposed in~\cite{peymanfard2023multi}. This model takes two separate audio and visual inputs and extracts the corresponding features using the trained AV-HuBERT model in the first step. We used the same pre-trained model for the audio part in both experiments. To ensure a fair comparison of the two trained models, we used an independent model for the audio part to isolate the effect of the two models. For the visual part, we used our proposed model and the model trained at the character level.

The first two rows of~\autoref{tab:av} show the comparison results of our proposed strategy with the baseline that uses characters in the fine-tuning stage. It is evident that our proposed method outperforms the baseline in both metrics. We also compared the results for two common decoding methods: greedy and beam search. As expected, the beam search achieves better results based on both CER and WER.

\begin{table*}[btp!]
  \renewcommand{\arraystretch}{1.2}
  \begin{center}
    \caption{Word-level Visual Speech Recognition Accuracy in Percentage}
    \label{tab:acc}
    \setlength\tabcolsep{8pt}
    \begin{tabular}{l | c c c c} 
      & \textbf{Accuracy} & \textbf{Top3 Acc.} & \textbf{Top5 Acc.} & \textbf{Top10 Acc.} \\
      \textbf{Baseline Model~\cite{Peymanfard2022Dataset}} & 24.79 & 38.79 & 45.47 & 55.03 \\
      \textbf{Fine-tuned by Characters} & 23.66 & 38.14 & 47.23 & 54.31 \\
      \textbf{Fine-tuned by Visemes} & 25.90 & 39.64 & 47.72 & 56.21 \\
    \end{tabular}
  \end{center}
\end{table*}

\begin{table*}[btp!]
  \renewcommand{\arraystretch}{1.2}
  \begin{center}
    \caption{Audio-Visual Speech Recognition Error Rates in Percentage. Greedy and Beam indicate the used decoding strategies. The external dataset also indicates the situation in that we used an external audio dataset in the training (see Section~\ref{sec:external}).}
    \label{tab:av}
    \setlength\tabcolsep{8pt}
    \begin{tabular}{l | cccc} 
     & \textbf{CER} & \textbf{CER} & \textbf{WER} & \textbf{WER} \\
      \textbf{Decoding Method} & Greedy & Beam & Greedy & Beam \\
      \textbf{Fine-tuned by Characters} & 14.56 & 13.84 & 46.63 & 45.12 \\
      \textbf{Fine-tuned by Visemes} & 13.56 & 12.87 & 44.18 & 42.63 \\
      \textbf{Fine-tuned by Characters (External dataset)} & 8.84 & 8.66 & 28.71 & 28.19 \\
      \textbf{Fine-tuned by Visemes (External dataset)} & 8.11 & 7.95 & 25.88 & 25.37 \\
    \end{tabular}
  \end{center}
\end{table*}

\subsection{Audio-visual Speech Recognition using External Audio Dataset}
\label{sec:external}


In this experiment, we used Common Voice as an external dataset in addition to Arman-AV. The Common Voice dataset contains over 352 hours of speech. The aim of this experiment is to examine the effect of the features extracted by our method on the model that extracts high-quality speech features. For this purpose, we trained the speech recognition model with the conformer-small network~\cite{gulati2020conformer} using both datasets. Then, we used the encoder part of the trained network to extract speech features for each sample in the Arman-AV dataset. Finally, we trained the audio-visual speech recognition model using the model described in the previous section. 

The next two rows of \autoref{tab:av} show the comparison results when an external audio dataset was used in the training. By comparing these results with the results of the previous section, we can see that using audio-only external datasets can greatly reduce errors. Also, here the proposed viseme-based strategy consistently outperforms the baseline.

\begin{figure*}[btp!]
    \centering
    \subfloat[The proposed method]{\includegraphics[width=0.5\textwidth]{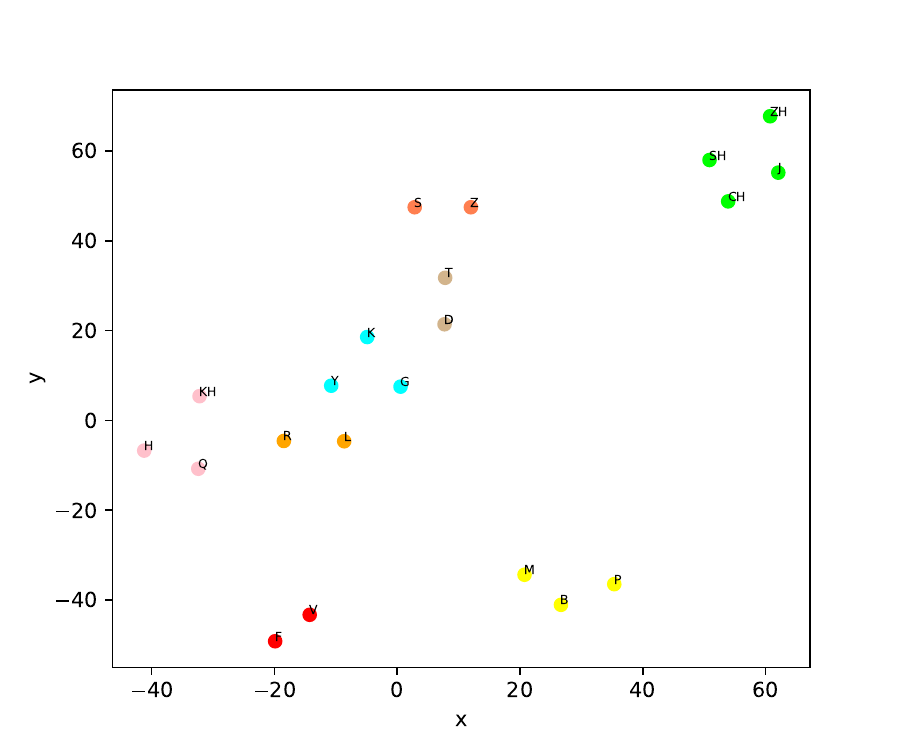}}
    \subfloat[AV-HuBERT baseline model]{\includegraphics[width=0.5\textwidth]{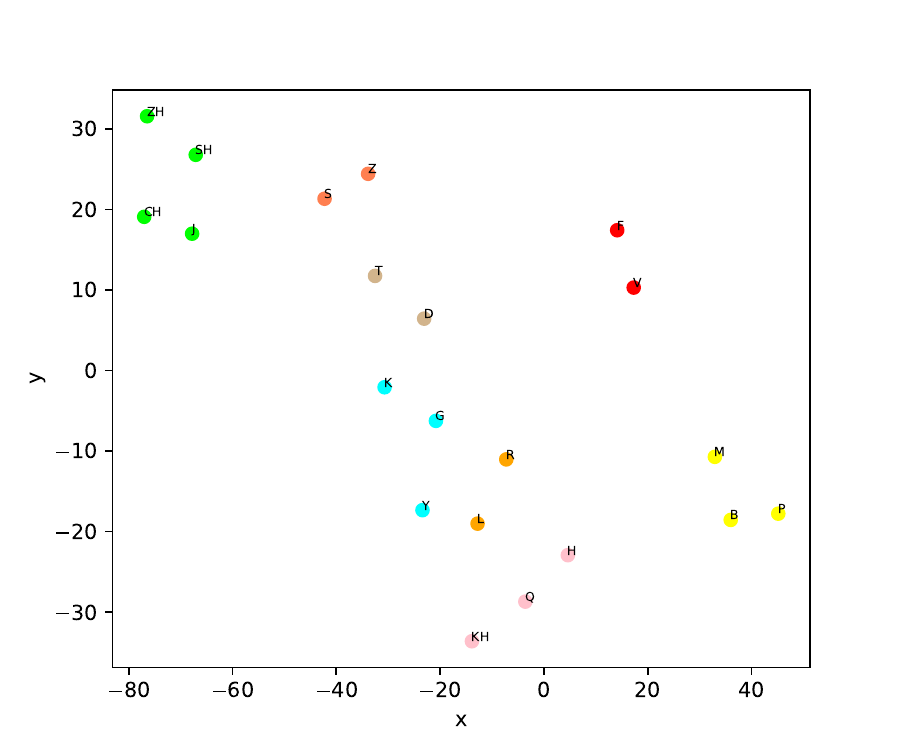}}
    \caption{2D representation of Persian phonemes using a visual representation model by averaging embeddings for each phoneme obtained from phoneme-level labeled data.}
    \label{fig:tsne}
\end{figure*}
\section{Analysis}
\label{sec:analysis}


In this section, we aim to compare the two models. To do so, we extract and analyze the embedding of each phoneme. We use the ARMAN-AV dataset for this purpose, which has phoneme-level transcription. For each phoneme, we compute the average visual embedding. We then analyze these embeddings using two methods. The first method involves mapping the 768-dimensional vector for each model to a two-dimensional space using t-SNE (\autoref{fig:tsne}). The second method involves clustering the embeddings and comparing the quality of the obtained clusters.


The embeddings of the phonemes are compared using a clustering evaluation metric called Silhouette. In this comparison, we first assign a cluster to each phoneme. Each phoneme belongs to a cluster that has similar lip movements to other phonemes in the same cluster. In fact, we group the phonemes of each viseme into a cluster. The proposed model improves the clustering quality for phonemes that have a high degree of visual similarity to each other, as shown in~\autoref{tab:sil1}. The results in this table show that our proposed model achieves higher-quality clustering of the video features for the lip-reading problem. This means that the phonemes that have similar lip movements are closer to each other in the feature space. This improvement is remarkable considering that the model was trained on English data, but the analysis was performed on Persian data.

\begin{table*}[btp!]
  \renewcommand{\arraystretch}{1.2}
  \begin{center}
    \caption{Silhouette Score Comparison}
    \label{tab:sil1}
    \setlength\tabcolsep{12pt}
    \begin{tabular}{l | cc} 
     & \textbf{Silhouette}\\
      \textbf{Pre-trained AV-HuBERT} & 33.79 \% \\
      \textbf{The Proposed Method} & 36.21 \% \\
    \end{tabular}
  \end{center}
\end{table*}
\section{Conclusion}

Lip reading is a challenging task that has received significant attention in recent years due to its potential applications in various fields. However, developing an accurate lip-reading system is still a difficult task due to the scarcity of data and the similarity of lip movements for some phonemes (these similar phonemes are called visemes). AV-HuBERT is a self-supervised method that can learn audio and visual features. We proposed a method to fine-tune the AV-HuBERT model using visemes to enhance the visual feature extraction. We evaluated our proposed method on the Arman-AV dataset, using different tasks such as word and sentence level lip reading, and audio-visual speech recognition. In all of them, our proposed method outperformed the baseline AV-HuBERT model. As future work, we plan to extend our approach to a multilingual setting.

~\\

\bibliographystyle{IEEEtran}
\bibliography{end/lipref}

\end{document}